\documentclass[10pt, a4paper]{article}

\usepackage{lrec-coling2024}

\usepackage{multirow}
\usepackage{graphicx}
\usepackage{subfigure}
\usepackage{amsfonts}
\usepackage{times}
\usepackage{latexsym}
\usepackage{amsmath}
\usepackage{amsfonts}
\usepackage{multirow}
\usepackage{booktabs,arydshln}
\usepackage{graphicx}
\usepackage{algorithm}
\usepackage{algorithmic}
\usepackage{tabu}
\usepackage{xcolor}
\usepackage{hyperref}
\usepackage{setspace}

\title{MulCogBench: A Multi-modal Cognitive Benchmark Dataset for Evaluating Chinese and English Computational Language Models}

\name{\normalsize{Yunhao Zhang\textsuperscript{\rm 1,2,†}, Xiaohan Zhang\textsuperscript{\rm 1,2,†}, Chong Li\textsuperscript{\rm 1,2,†}, Shaonan Wang\textsuperscript{\rm 1,2}, Chengqing Zong\textsuperscript{\rm 1,2}}}

\address{\textsuperscript{\rm 1}State Key Laboratory of Multimodal Artificial Intelligence Systems, Institute of Automation, CAS\\
\textsuperscript{\rm 2} School of Artificial Intelligence, University of Chinese Academy of Sciences \\
         \{zhangyunhao2021, lichong2021\}@ia.ac.cn \\
         \{xiaohan.zhang, shaonan.wang, cqzong\}@nlpr.ia.ac.cn\\}

\abstract{
Pre-trained computational language models have recently made remarkable progress in harnessing the language abilities which were considered unique to humans. Their success has raised interest in whether these models represent and process language like humans.
To answer this question, this paper proposes MulCogBench, a multi-modal cognitive benchmark dataset collected from native Chinese and English participants. It encompasses a variety of cognitive data, including subjective semantic ratings, eye-tracking, functional magnetic resonance imaging (fMRI), and magnetoencephalography (MEG). 
To assess the relationship between language models and cognitive data, we conducted a similarity-encoding analysis which decodes cognitive data based on its pattern similarity with textual embeddings.
Results show that language models share significant similarities with human cognitive data and the similarity patterns are modulated by the data modality and stimuli complexity. Specifically, context-aware models outperform context-independent models as language stimulus complexity increases.
The shallow layers of context-aware models are better aligned with the high-temporal-resolution MEG signals whereas the deeper layers show more similarity with the high-spatial-resolution fMRI. These results indicate that language models have a delicate relationship with brain language representations. 
Moreover, the results between Chinese and English are highly consistent, suggesting the generalizability of these findings across languages.
 \\ \newline  }

\begin{document}

\maketitleabstract
\def\thefootnote{†}\footnotetext{Equal Contribution}\def\thefootnote{\arabic{footnote}}
\section{Introduction}

The ability to use language has long been considered unique to the human species. 
However, the emerging pre-trained language models have achieved super-human performance on various language tasks \citep{Zhao2023ASO,Huang2022TowardsRI,Bang2023AMM}. 
Their success has elicited a flurry of research and debates about the following question: how similar are the working mechanisms of these language models to that of the human brain \citep{Blank2023WhatAL}? To answer this question, high-quality cognitive data collected while human participants are understanding language is indispensable.

Human cognitive data, including brain activation and behavioral reaction, reflects how the human brain processes language.
There has been evidence that the representations of computational language models, even though trained solely on large text corpora, share similarities with the brain activation elicited by language \citep{gauthier-levy-2019-linking,hashemzadeh-etal-2020-language,pasquiou2022neural} . Moreover, such similarity pattern is affected by factors including model architectures, loss functions and etc. \citep{caucheteux2022brains}. 
These findings suggest that language models may mimic the human brain in language processing to different degrees. 
The more similar the representations of language models to human cognitive data, the more possibility there is that the two systems share the same mechanisms.
Therefore, cognitive data not only plays an important role in studying the brain mechanism of language but also can serve as criteria for the cognitive plausibility of computational models. 


There have been a few cognitive benchmark datasets \cite{anderson-etal-2013-words,xu-etal-2016-brainbench,hollenstein-etal-2019-cognival}. However, the available datasets are either too small or only focus on English. Moreover, evaluating computational models using cognitive data is still rarely done outside psycholinguistics and neurolinguistics. Accordingly, it is still unknown 1) 
whether computational models have their better-aligned cognitive modality, 
2) whether their mechanisms in processing different linguistic units are similar to that of humans, and 3) whether the relationship between computational models and cognitive data can be generalized to different languages.

To answer the above questions, this paper presents MulCogBench, a multi-modal cognitive benchmark dataset for evaluating Chinese and English language models. MulCogBench-Chinese encompasses cognitive data in four different modalities collected from Chinese native speakers, including behavioral data (here, word semantic rating and eye-tracking) and brain imaging data (here, fMRI and MEG). The language stimuli to collect these data ranges from words to discourses. MulCogBench-English involves three modalities, i.e., word semantic rating, eye-tracking, and fMRI. To evaluate the similarities between the representations of computational models and the human brain, we conduct experiments on four classic computational language models, namely Word2vec, GloVe, BERT, and GPT-2. Specifically, we employed similarity-encoding analysis to compute the representational similarities between language models and cognitive data. In particular, for the high-spatial-resolution fMRI, we performed a fine-grained ROI (region of interest) level analysis based on the functional division of the brain.

Results show that computational models have significant similarity with human cognitive data, in which the similarity degree is modulated by cognitive modalities and linguistic units. Specifically, we find that (1) across different cognitive modalities and linguistic units, the most-similar models and the variation tendencies within the context-aware models are highly consistent between Chinese and English; (2) from word to discourse, the advantage of context-aware models over context-independent models increases, suggesting that these models are more human-like in encoding the complex language structure information but not in encoding the basic word-level information; (3) the similarity patterns between models and cognitive data vary in different cognitive modalities, with the shallow layers of context-aware models are more similar to MEG and the deeper layers are better aligned with fMRI, indicating that different layers may simulate different aspects of the human language mechanism.
Our results demonstrate that exploring the relationship between computational models and human cognitive data can help to explain the mechanism of both the computational model and the human brain.

All the cognitive data in MulCogBench will be released in the form that can be directly used for the computational model evaluation.





\begin{table*}[]
    \footnotesize
    \centering
    \begin{tabular}{c|c|c|c|c|c} \hline
       & \textbf{Modality} & \textbf{Source} & \textbf{Stimuli} & \textbf{Unit} & \textbf{Tokens} \\ \hline
       \multirow{5}{*}{Chinese} &word semantic rating & CRSF \cite{Wang2022AnFD} & text & word & 672 \\ \cline{2-6}
       &word fMRI & CRSF \cite{Wang2022AnFD} & text & word & 672 \\ \cline{2-6}
       & eye-tracking & \citet{zhang2022eyetracking} & text & sentence & 170,331 \\ \cline{2-6}
       &discourse fMRI & SMN4Lang \cite{Wang2022ASM} & audio & discourse & 52,269 \\ \cline{2-6}
       &discourse MEG & SMN4Lang \cite{Wang2022ASM} & audio & discourse & 52,269 \\ \hline
       \multirow{4}{*}{English} &word semantic rating& \citet{binder2016toward} & text & word & 535\\ \cline{2-6}
       & word fMRI & \citet{pereira2018toward} & text & word & 180\\ \cline{2-6}
       & eye-tracking & ZuCo \citep{hollenstein2018zuco,hollenstein-etal-2020-zuco} & text & sentence & 36,767 \\ \cline{2-6}
       & discourse fMRI &  \citet{zhang2020connecting} & audio & discourse & 47,356 \\ \hline
    \end{tabular}
    \caption{Details of the datasets in MulCogBench.}
    \label{tab:cogdata}
\end{table*}

\section{Cognitive Data}
Here we describe the sources of Chinese and English cognitive data in MulCogBench and how to use them. The MulCogBench includes cognitive modalities in eye-tracking, word semantic ratings, word fMRI, discourse fMRI, and MEG (only available for Chinese). See Table \ref{tab:cogdata} for more details. All the cognitive data were collected under the approval of the Institutional Review Board.

\subsection{MulCogBench-Chinese}
\paragraph{Eye-tracking} 
Eye-tacking records fine-grained temporal eye movement during reading, which provides information to study the cognitive mechanisms underlying reading \citep{rayner1998eye, rayner2009eye}. 
For example, early syntactic processing and lexical access are captured by early gaze measures when the first time a word is fixated.

We adopt a Chinese eye-tacking database obtained from 1,718 participants across 57 reading experiments \citep{zhang2022eyetracking}. 
It contains 7,577 natural Chinese sentences and 8,551 different words. 
The sentences involved range in length from 15 to 35, with an average length of 22.48.
There are nine word-level eye-tacking features provided: First Fixation Duration (FFD), Gaze Duration (GD), First-Pass reading Fixated proportion (FPF), Fixation Number (FN), proportion Regression In (RI), proportion Regression Out (RO), saccade length toward the target from the left (LI\_left), saccade length from the target to the right (LO\_right), and Total fixation duration (TT). 
All features are used in this study for a comprehensive investigation of the similarity between the human brain and language models. 

\paragraph{Word semantic ratings} 
To study how the brain represents word meaning, an essential start is to define a set of basic semantic features \citep{binder2016toward}. To evaluate whether the computational models represent word semantics in a similar way to that of humans, we adopt a Chinese semantic rating dataset that utilizes 54 semantic features defined from the functional division of the human brain, comprising both perceptual features, like vision and motor, and more abstract features, like social and emotion \citep{Wang2022AnFD}. This dataset includes 54 semantic feature ratings for 672 words. Each semantic feature of a word was annotated from a 1-7 score (score 1 means the feature is least associated with the word, and score 7 means the highest association) by 30 Chinese participants and averaged across all participants as the final rating score.


\paragraph{Word fMRI} 
As a noninvasive technique, fMRI measures the blood-oxygen-level-dependent (BOLD) signals, which reflect the neural activation through the blood flow changes. Here we use a Chinese fMRI dataset \cite{Wang2022AnFD} collected from 11 participants when they were reading the same 672 words as in the above word semantic ratings. During fMRI collection, each word was presented 6 times to participants with a unique corresponding image for each time. The fMRI data were pre-processed using fMRIPrep \cite{esteban2019fmriprep} and we conduct first-level analysis to obtain the neural activation of each word.

\paragraph{Discourse fMRI}
The discourse fMRI data was collected from 12 Chinese native speakers when they were listening to 60 stories \cite{Wang2022ASM}. Each participant listened to all 60 stories, and each story was listened to once by each participant. During the fMRI collection, participants were instructed to stay still and listen carefully to the stimuli. Each stimuli story last from 4 to 7 minutes. The total number of words in these stories is 52,269, and the vocabulary size is 9,153. The collected fMRI data were preprocessed following the HCP pipeline \cite{glasser2013the}.

\paragraph{Discourse MEG}
MEG is also a noninvasive brain mapping technique but measures magnetic fields produced by the electrical activity of neurons.
Different with fMRI, the MEG data has a relatively lower spatial resolution, typically with 300 or more sensors covering the head, but a high temporal resolution (millisecond timescale). The MEG data we use was collected from the same 12 participants as in the fMRI data described above \cite{Wang2022ASM}. The language stimuli were the same 60 stories as in the fMRI collection. There are 306 sensors in the MEG data. For each word, we calculate its MEG response by averaging its MEG signal in a 200 ms-long sliding window within 1 second after the word offset. Each time the window moves 100 ms backward, resulting in 9 response windows for each word.

\begin{figure*}[t]
\centering 
\setlength{\textfloatsep}{2pt}
\setlength{\intextsep}{2pt}
\setlength{\abovecaptionskip}{2pt}
\includegraphics[width=0.8\textwidth]{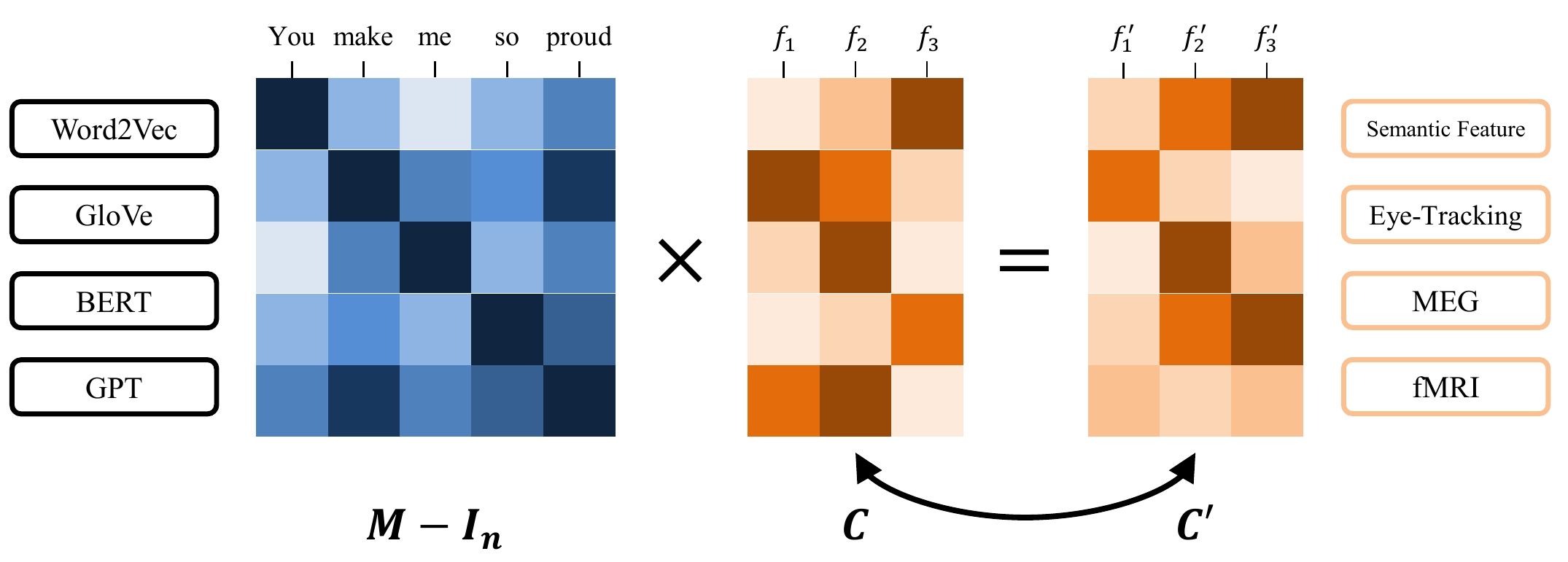}
\caption{The procedure of similarity-encoding analysis (SEA). The representation similarity matrix $M$ is calculated by the embeddings from computational language models. }\label{fig:sea}
\end{figure*}

\subsection{MulCogBench-English}

\paragraph{Eye-tracking} 
ZuCo 1.0 and 2.0 eye-tracking databases are used as English eye-tracking data \citep{hollenstein2018zuco,hollenstein-etal-2020-zuco}. 
There are two reading tasks, the normal reading and the task-specific reading task, from 30 subjects across experiments. 
To better align with the input of computational language model, which is not used to solve specific tasks, only the eye-tracking data in the normal reading task are preserved in our experiments, which contain 1,049 sentences. 
Six word-level eye-tracking features are extracted from the raw eye-tracking recordings: Gaze Duration (GD), Total Reading Time (TRT), First Fixation Duration (FFD), Single Fixation Duration (SFD), Go-Past Time (GPT), and the number of fixations (nFix).

\paragraph{Word semantic ratings}
The English semantic dataset we use was proposed by \newcite{binder2016toward}, which included 535 concepts and each concept has 65 semantic features. Same as in Chinese semantic rating data, each semantic feature of a word is the averaged score of 30 participants. Differently, this dataset was annotated with a saliency score on a 0-6 scale.


\paragraph{Word fMRI}
We use the English word fMRI dataset that contains 180 words involving 131 nouns, 22 verbs, 21 adjectives, and 6 adverbs\cite{pereira2018toward}. This dataset was collected from 15 native English speakers when they were reading the word with a corresponding image. Each participant saw a word between 4 and 6 repetitions with a unique picture at each time.
After preprocessing, first-level analysis was conducted to acquire the neural activation of each word.

\paragraph{Discourse fMRI}
The English discourse fMRI was collected from 19 native English speakers when they were listening to 51 English stories. Each participant listened to a subset of stories and each story lasts from 4 to 13 minutes. In total, the
story stimuli include 47,356 words, forming a vocabulary of 5,228 words. The fMRI data was also preprocessed following the HCP pipeline \cite{glasser2013the}. 

\section{Computational Language Models}
We adopt the four most commonly-used computational language models that can be divided into two groups: one is context-independent model including Word2Vec \citep{mikolov2013cbow} and GloVe \citep{pennington-etal-2014-glove}, which calculate word representations based on word occurrence in raw texts. The other is the context-aware model including BERT \citep{cui-etal-2020-revisiting, devlin-etal-2019-bert} and GPT family models \citep{radford2019gpt2}. Among these, BERT is an autoencoder language model, trained bidirectionally to predict masked tokens. GPT is an autoregressive language model, trained to predict the next token based on preceding text.


For Chinese, the Word2vec and GloVe embeddings were both trained on the Xinhua News corpus (19.7 GB)\footnote{http://www.xinhuanet.com/whxw.htm} with the same model parameters (i.e., Skip-Gram architecture, negative number as 15 in Word2Vec, window width of 2, embedding dimensions of 300). For English, the Word2vec and GloVe embeddings were trained with the Wikipedia corpus (13 GB)\footnote{https://dumps.wikimedia.org/enwiki/latest} with the same model parameters as Chinese models.
Pre-trained MacBERT\footnote{https://huggingface.co/hfl/chinese-macbert-base} and GPT-2\footnote{https://huggingface.co/uer/gpt2-chinese-cluecorpussmall} models for Chinese and the BERT\footnote{https://huggingface.co/bert-base-uncased} and GPT-2\footnote{https://huggingface.co/gpt2} for English were downloaded from HuggingFace. Both the BERT-based and GPT-based models have 12 hidden layers, which were all used in the experiments. 
See more detailed model parameters in Table \ref{tab:embed}. 

\begin{table}[h]
    \footnotesize
    \centering
    \begin{tabular}{c|c|c|c} \hline
        & \textbf{Model} & \textbf{Dim} & \textbf{Layers} \\ \hline
        \multirow{4}{*}{Chinese} & Word2Vec & 300 & 1 \\ \cline{2-4} 
        & GloVe & 300 & 1 \\ \cline{2-4} 
        & MacBERT & 768 & 12 \\ \cline{2-4} 
        & GPT-2 & 768 & 12 \\ \hline
        \multirow{4}{*}{English} & Word2Vec & 300 & 1 \\ \cline{2-4} 
        & GloVe & 300 & 1 \\ \cline{2-4} 
        & BERT-base-uncased & 768 & 12 \\ \cline{2-4} 
        & GPT-2 & 768 & 12 \\ \hline
    \end{tabular}
    \caption{Parameters of computational language models.}
    \label{tab:embed}
\end{table}

\section{Evaluation Methods}
To evaluate how human-like the computational language models are, we conducted similarity-encoding analysis (SEA), which reconstructs the cognitive data $C$ based on the representational similarity between $C$ and textual embeddings $E$ of models (Figure \ref{fig:sea}). Specifically, the SEA computes the similarity between the embeddings of the words in $E$ and reconstructs the corresponding cognitive data of each word by adding the cognitive data of other words weighted by the computed embedding similarity. The motivation is that if the computational model encodes linguistic information like humans, then the similarity patterns between textual embeddings and the cognitive data should be similar. Therefore, the reconstructed cognitive data would be more similar to the original data. The procedures of SEA are described as follows:


First of all, we have a group of word or sentence embeddings $E=\{e_1,e_2,...,e_n\}$ computed by different models. For each pair of embedding $(e_i,e_j)$, its similarity is measured by the Pearson correlation coefficient ($\rho$). 
Thus we have a similarity matrix $M\in \mathbb{R}^{n\times n}$, where $M_{ij}=\rho (e_i,e_j)$.



\indent Then, for the cognitive data, we assumed that if a specific cognitive representation encodes the same information as embeddings, the similarity relation of these embeddings and that of the cognitive vectors was the same. 
We can predict each cognitive vector by multiplying the above similarity matrix with corresponding cognitive vectors. In addition, to remove the influence of the ground truth value, we subtracted the real cognitive vectors from the predicted cognitive vectors to obtain the predicted cognitive matrix.
\begin{equation}
    C^{\prime}= (M-I_n)C
\end{equation}
where $C\in \mathbb{R}^{n\times m}$ is the real cognitive vectors, and $C^{\prime}\in \mathbb{R}^{n\times m}$ is the predicted cognitive vectors.

Finally, the Pearson correlation was calculated to evaluate the similarity between the predicted and the real cognitive vectors: 
\begin{equation}
    r=\frac{1}{n} \sum_{i=1}^n \rho (C_{i,:}, C^{\prime}_{i,:})
\end{equation}
A higher correlation score $r$ means that the information in the cognitive data is better encoded in the specific computational model. 

The procedure to conduct SEA is slightly different for each modality of cognitive data due to their unique property.

\paragraph{Eye-tracking}
For the eye-tracking features, we concatenated all sentences together and conducted SEA for each eye-tracking feature. 



\paragraph{Word semantic rating}
For 54 features, we computed correlation for each feature and averaged across features as the final result.

\paragraph{Word fMRI}
For word fMRI, we have one brain activation vector for each word. We conducted an ROI-level analysis (the ROIs were adopted following \citet{beam2021data}).
To select the most informative voxels in each ROI, we trained regression models for each voxel to predict each the word embeddings with this voxel and its 26 adjacent 3D neighbors. The correlation between the true and the predicted embeddings was computed and then as the informativeness score of each voxel. The top 10\% voxels with the highest scores within each ROI were chosen for SEA analysis.

\paragraph{Discourse fMRI}
For discourse fMRI, we also conducted the ROI-level anaylsis as in word fMRI and used voxel-wise encoding to choose highly informative voxels in each ROI.
Because the change of BOLD signal lasts for tens of seconds after the neurons fire and the temporal resolution of fMRI is comparatively lower than words, we first convolved the word embeddingss with the canonical hemodynamic response function (HRF)\footnote{The canonical HRF describes how BOLD signals changes after the neurons fire.} and downsampled the convolved features to the sampling rate of fMRI. Then, for each voxel, we trained a linear regression model to predict its response with downsampled features. Finally, the Pearson correlation was computed between predicted and actual fMRI data, and the top 10\% voxels with the highest correlation were chosen.

\paragraph{Discourse MEG}
To choose the most informative sensors and time windows in MEG, we trained linear regression models to predict the signal of each sensor at each sliding window using word embeddings. In the experiments, we only show the top 5\% sensors and the time window that achieves the highest prediction accuracy.

\begin{figure*}[ht]
    \flushleft
    \subfigure[]{\includegraphics [scale=0.3] {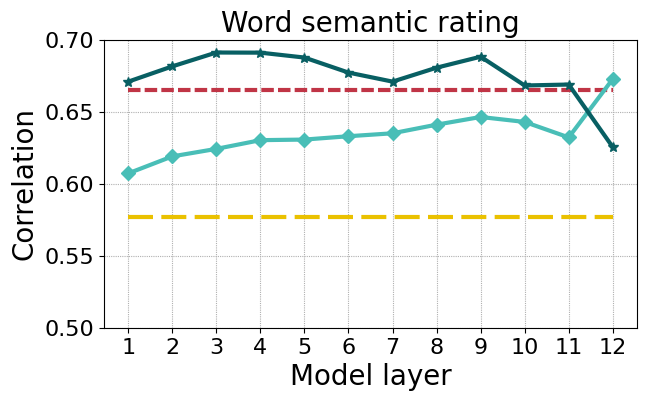}}
    \subfigure[]{\includegraphics [scale=0.3]{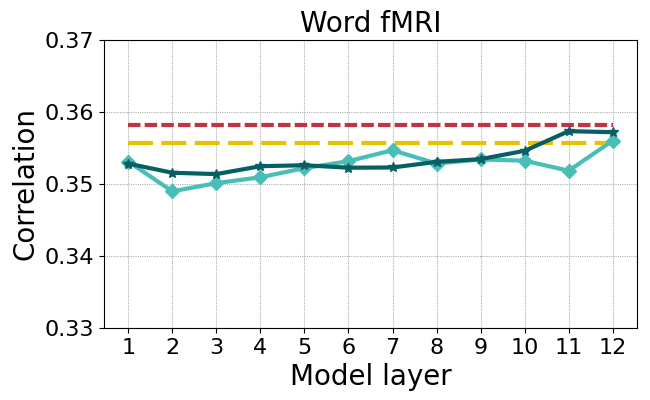}}
    \subfigure[]{\includegraphics[scale=0.3]{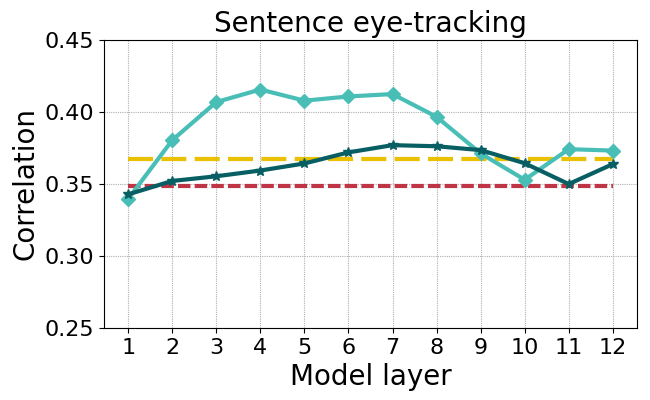}}
    \subfigure[]{\includegraphics[scale=0.3]{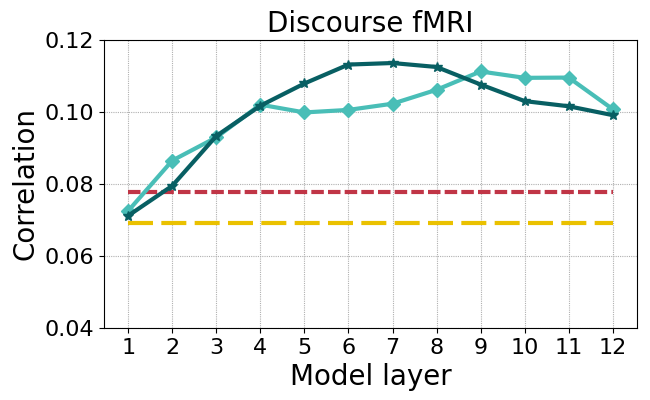}}
    \subfigure[]{\includegraphics[scale=0.3]{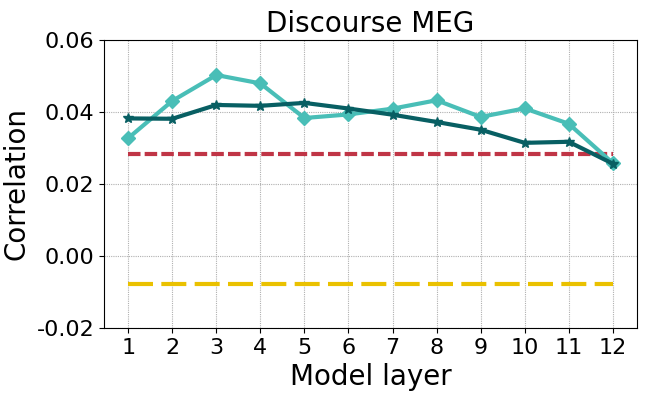}}
    \subfigure{\includegraphics[scale=0.5]{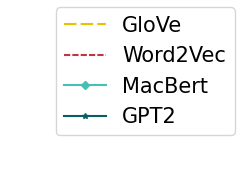}}
    \caption{Evaluation Results on MulCogBench-Chinese dataset.}
    \label{fig:results_avg_zh}
\end{figure*}

\begin{figure*}[ht]
    \flushleft
    \subfigure[]{\includegraphics [scale=0.3] {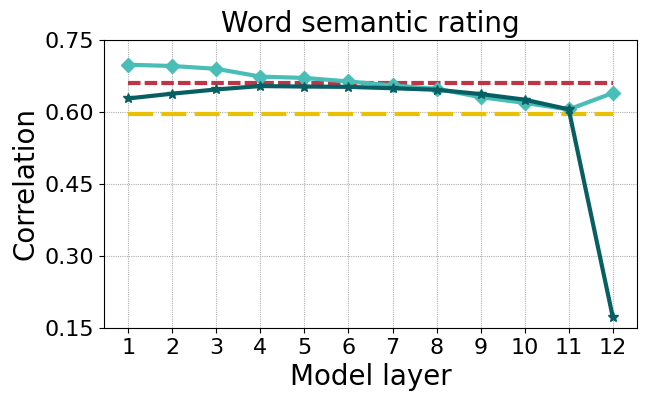}}
    \subfigure[]{\includegraphics [scale=0.3]{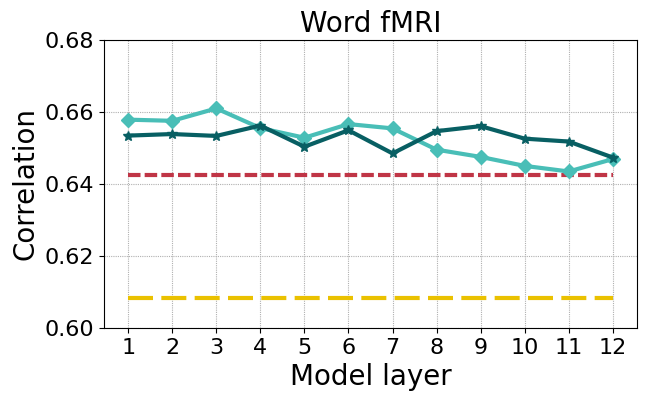}}
    \subfigure[]{\includegraphics[scale=0.3]{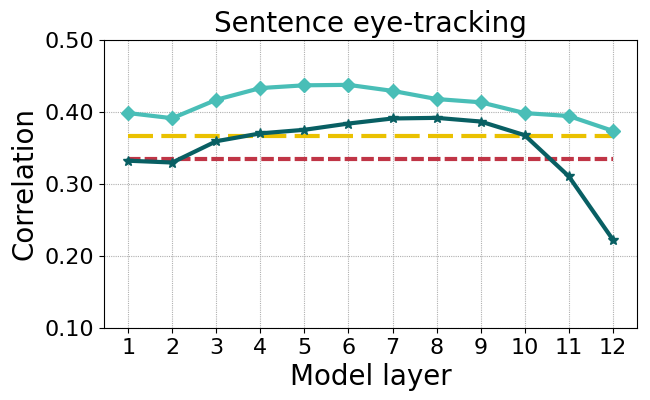}}
    \subfigure[]{\includegraphics[scale=0.3]{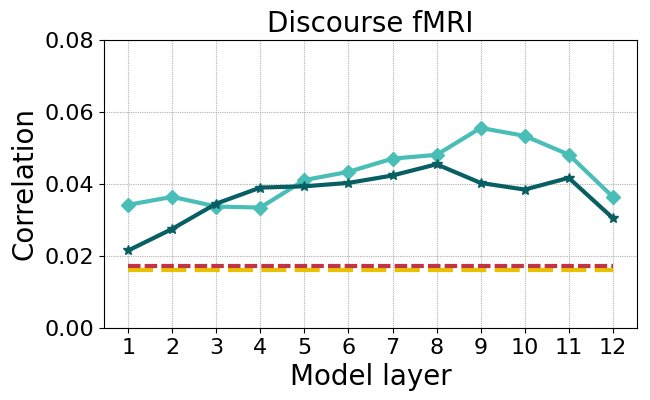}}
    \subfigure{\includegraphics[scale=0.5]{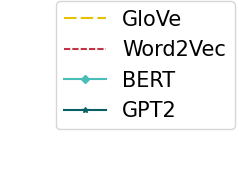}}
    \caption{Evaluation Results on MulCogBench-English dataset.}
    \label{fig:results_avg_en}
\end{figure*}

\begin{figure*}[ht]
    \centering
    \subfigure[Chinese word fMRI]{\includegraphics [scale=0.57] {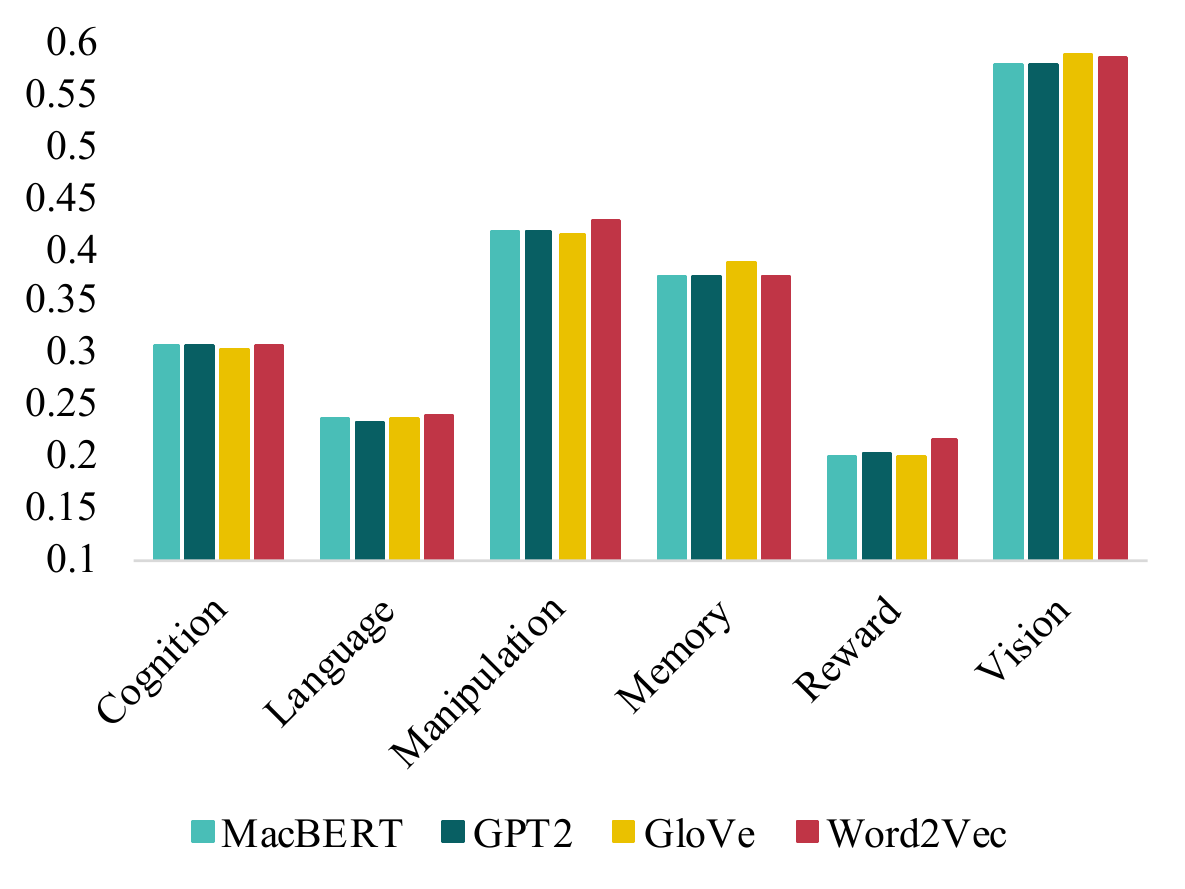}}
    \subfigure[Chinese discourse fMRI]{\includegraphics[scale=0.57]{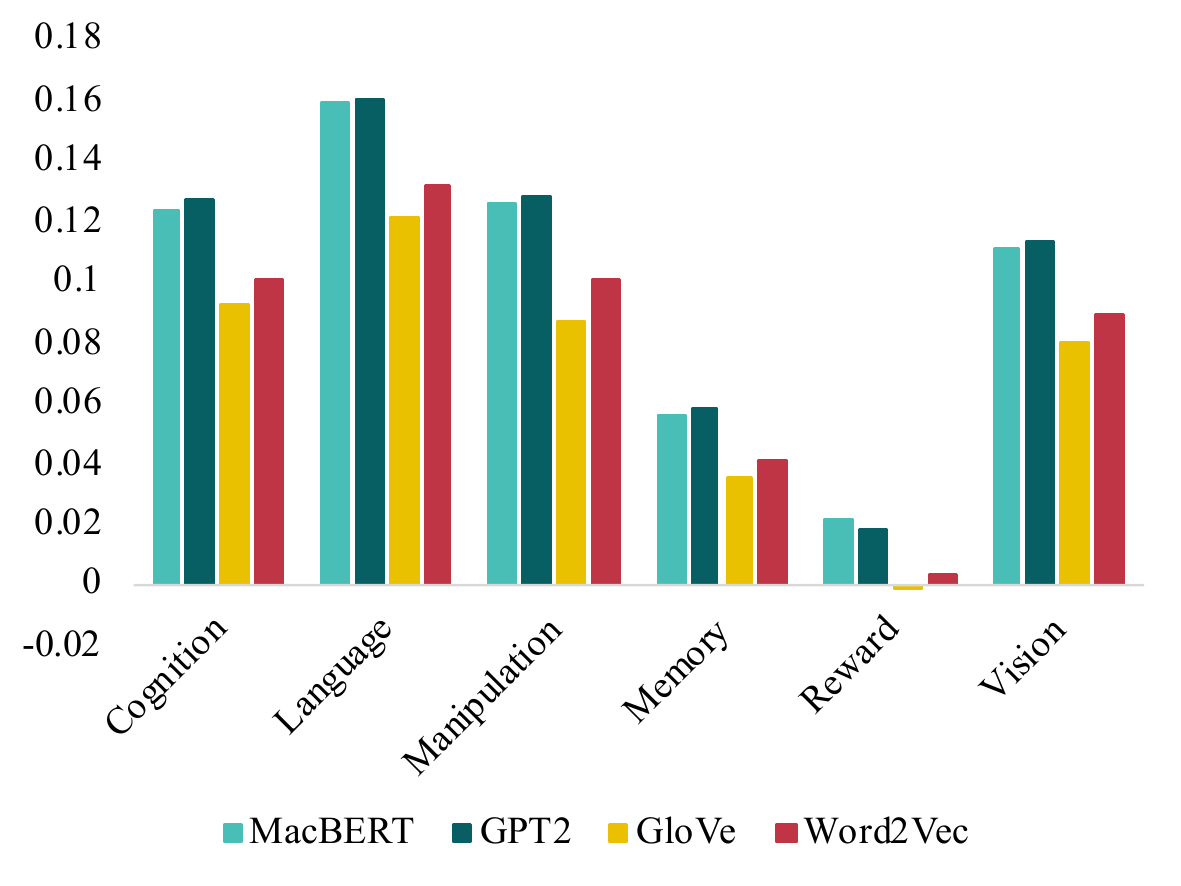}}
    \subfigure[English word fMRI]{\includegraphics [scale=0.57]{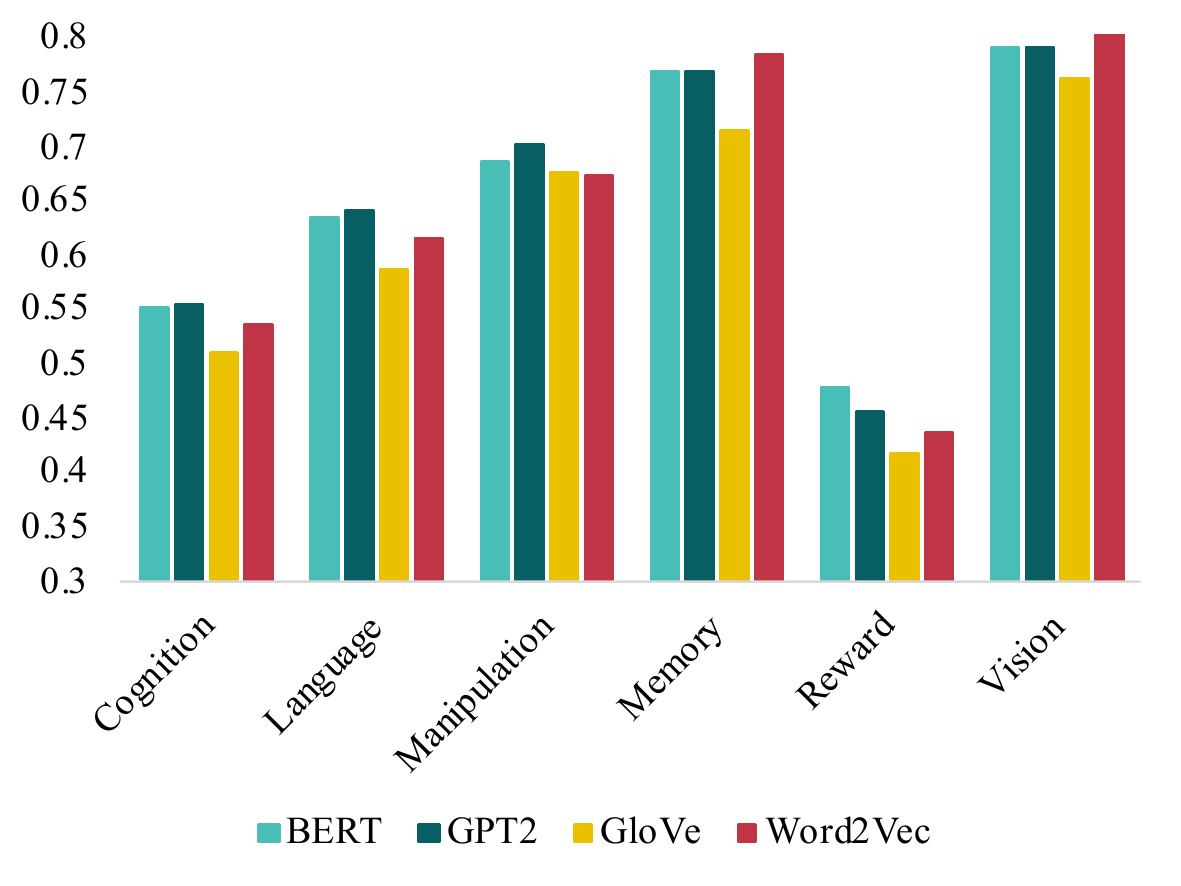}}
    \subfigure[English discourse fMRI]{\includegraphics[scale=0.57]{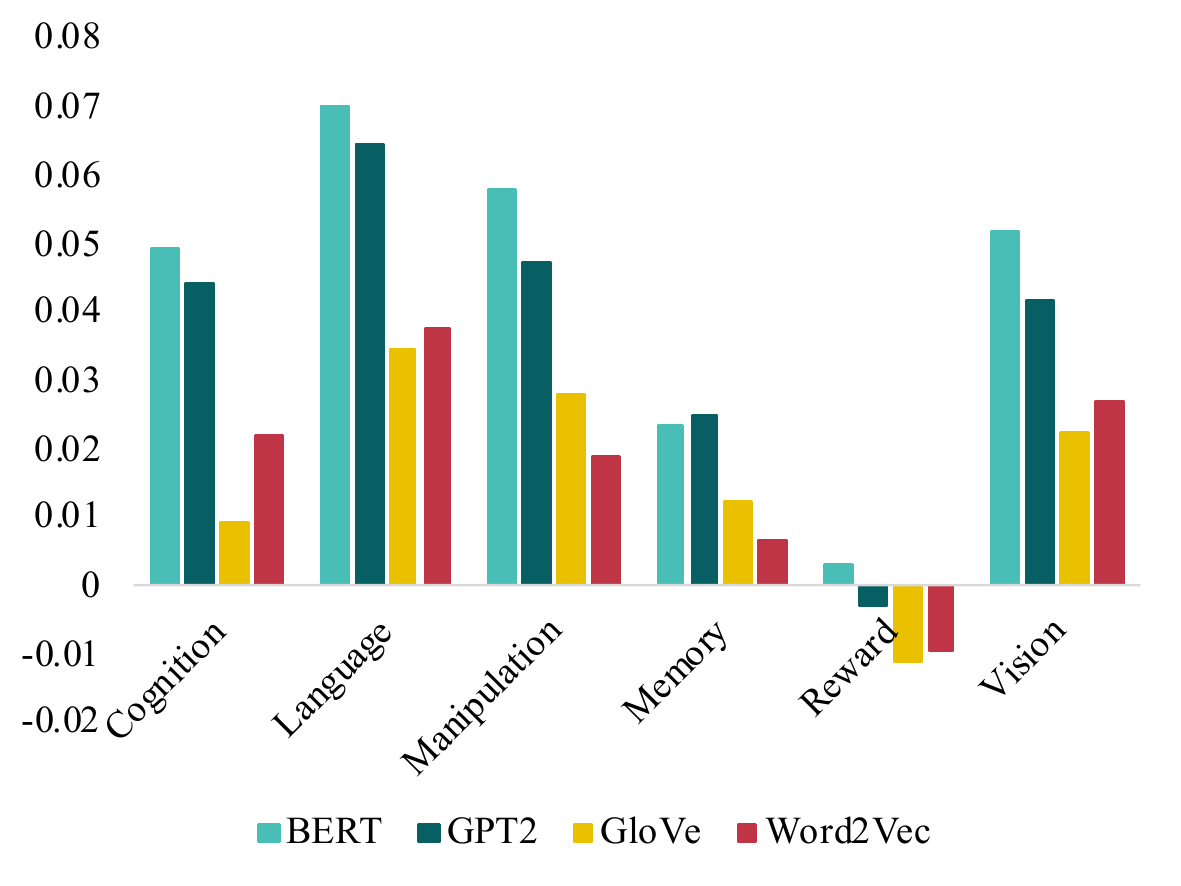}}
    \caption{Evaluation results on different ROIs of fMRI.}
    \label{fig:results_roi}
\end{figure*}


\section{Results and Analysis}
Figure \ref{fig:results_avg_zh} and Figure \ref{fig:results_avg_en} illustrate the evaluation results of each cognitive modality.
For cognitive modalities with multiple features, we average the correlations across all features as the final result.
As shown in most cognitive modalities, the SEA results are significantly larger than the random result, which is 0, indicating that computational models share significant similarities with the cognitive data.
However, the similarity patterns are different across cognitive modalities and linguistic units.
We analyze how the cognitive modality and the linguistic unit modulate the similarity between language models and the cognitive data in the following.

\paragraph{Comparison between Chinese and English}
As shown in Figure \ref{fig:results_avg_zh} and Figure \ref{fig:results_avg_en}, for both Chinese and English, computational language models show significant correlation with cognitive data. More importantly, the similarity patterns between these two languages are highly consistent across cognitive modalities and linguistic units. Specifically, the best-performed models and the variation tendency from shallower to deeper layers within context-aware models are very similar between Chinese and English. In word-level fMRI, the correlation for both languages show no significant difference\footnote{The significance level mentioned in this section is $\alpha=0.001$} between any two layers of context-aware models or between any layers of context-aware models and the context-independent models. The only exception is that context-aware models significantly outperform GloVe in English. Moreover, as the complexity of linguistic units increases, the consistency between Chinese and English becomes more clear.
Since Chinese and English are two diverse languages, these consistencies suggest that the relationship between computational models and cognitive data is largely generalizable across languages, at least between Chinese and English. Therefore, the following analyses are conducted based on these consistencies.

Although the results of these two languages are consistent in most circumstances, the correlation of GPT-2 drop dramatically at 12th layer in English but not in Chinese. We have conducted multiple checks to ensure the its accuracy and have identified two possible reasons for this phenomenon.Firstly, most dimensions in GPT-2 embeddings have values ranging from -2 to 2, but several dimensions in the 12th layer of the English GPT-2 model have extremely large values, even exceeding 200. This is markedly different from the other 11 layers and was not observed in the 12th layer of the Chinese GPT-2 model.Secondly, our semantic rating score is an average of 65 semantic features, and the substantial drop was only observed in a specific subset of these features. This suggests that this subset of features may be particularly sensitive to the abnormal dimensions in the 12th layer of the GPT-2 model.

\paragraph{Comparison between linguistic units}
The most critical property of human language is its combinatorial nature \cite{Ding2015CorticalTO}. When combining smaller elements such as words into larger structures like sentences, the involved cognitive processes are different.
Accordingly, the similarity patterns between computational models and cognitive data also vary in different linguistic units as illustrated in Figure \ref{fig:results_avg_zh} and Figure \ref{fig:results_avg_en}.

In word-level, although the exact patterns differ between the two modalities, i.e., semantic rating and fMRI, a unique phenomenon is that the context-aware models do not outperform context-independent models in most cases. In word fMRI, there is no significant difference between these two types of models except that the English context-aware models significantly outperform GloVe. In word semantic rating, although GloVe has the worst performance, the resulting correlation of Word2Vec is significantly higher than the deep layers of both two context-aware models. 
In contrast to word-level results, context-aware models have a significant advantage over context-independent models in sentence-level and discourse-level. Many layers of BERT/MacBERT and GPT-2 reached higher correlation than Word2Vec and GloVe.
This advantage is especially significant in discourse-level, in which all except the first layer of BERT/MacBERT and GPT-2 have a higher correlation than GloVe and Word2Vec. 
These results suggest that context-aware models only become more human-like than context-independent models when processing complex linguistic units.

Apart from the differences between context-aware and context-independent models, the variation tendency from shallower to deeper layers within context-aware models also differs in three linguistic units. In word semantic rating, the shallow layers of BERT/MacBERT and GPT-2 have higher performance than the deep layers. Whereas in sentence and discourse fMRI, the middle and deep layers reached the highest correlation. 
These layer-differences indicate that as the layer goes deeper in context-aware models, the encoded information transfers from simple to more complex linguistic units.


\paragraph{Comparison between the modalities of cognitive data}
Both fMRI and MEG are direct measurements of the brain signals, whereas semantic rating and eye-tracking are behavioral reactions. These modalities of cognitive data reflect different aspects of the human language mechanism. Therefore, comparing the performance of models in different cognitive modalities is helpful for further understanding how these models are similar to the human language mechanism. Since the model performance is modulated by the complexity of linguistic unit, we only analyze the modality differences within the same linguistic unit, that is, we compare the results between fMRI and semantic rating in word-level and fMRI and MEG in discourse-level. 

In word-level, the significant difference between models was only found in semantic rating modality which reflects how human understand the word semantics through the extrinsic and subjective rating behavior. Whereas the measured fMRI is an objective signal of how the brain respond the words. A possible reason for this modality-difference may be that decoding fMRI is more difficult than decoding semantic features from embeddings. The model-differences in encoding word-level information is concealed by the difficulty in decoding fMRI.
In discourse-level, the middle and deep layers better predict fMRI while shallow layers better predict MEG. Since fMRI is a high-spatial resolution signal, a better-performed model in fMRI is more likely to be similar to brain in the spatial activation patterns. Whereas MEG is a high-temporal resolution signal, a better-performed model is more likely to better encode the rapid word changes.

These variations in similarity patterns in modalities suggest that different layers of models may simulate different aspects of human language mechanisms. Further analysis of these variations may be helpful for improving the model to process language with higher efficiency and accuracy.

\paragraph{Comparison between ROIs}
Since the variation tendency within context-aware models is similar to that in Figure \ref{fig:results_avg_zh} and Figure \ref{fig:results_avg_en}, we averaged the results of all layers within BERT/MacBERT and GPT-2 and focus on the overall performance of models in each ROI.
As shown in Figure \ref{fig:results_roi}, the performance of computational models in each ROI is similar to the ROI-averaged results. The context-aware models only outperform context-independent models in discourse-level. Besides, the performance difference between models is consistent in all 6 ROIs. However, the most related ROI with computational models changes from Vison to Language when the linguistic unit changes from word to discourse.
We do not attribute this change to the brain mechanism shift between linguistic units. Rather, the presentation form of stimuli may highly correlate with this phenomenon. In word-level fMRI collection, participants watched the words and the related pictures, while in discourse fMRI collection,  the stories were played to participants with a blank screen showing nothing but a plus symbol.

Apart from the Vision and Language, brain regions involved in manipulation, cognition, and memory also have significant correlation with computational models. These results suggest that the brain foundation for language may be broader than traditional language brain networks.

\section{Conclusion and Future Work}
How similar the working mechanism of computational language models is to that of the human brain has raised great interest.
This paper provides the MulCogBench dataset which consists of rich cognitive data that can be used to explore this question in many aspects.
Specifically, this paper investigated their representational similarity from three aspects, i.e., the modality of cognitive data, the linguistic unit, and the cross-linguistic commonality. We find that both the modality of cognitive data and the linguistic unit affect the similarity patterns between computational models and the cognitive data.

As a large-scale cognitive dataset, MulCogBench includes multiple features within each modality which can be used for fine-grained analysis. For instance, word semantic ratings can be used to explain what semantic features are encoded in textual embeddings and to explore whether these features in textual embeddings are organized in a similar way as encoded in the brain. Testing the similarity is the first step in exploring the underlying cognitive mechanisms. Hopefully, the MulCogBench will contribute to a better understanding of both the language mechanism of computational models and the human brain.

\section{Limitations}
Cognitive data including eye-tracking and neuroimaging such as fMRI and MEG contains a lot of noise and each represents multiple cognitive functions. For instance, the first-time passes of words in eye-tracking not only reflect the attention given to the words but also the visual recognition and other word-level or sentence-level processes. Therefore, it is difficult to directly correlate a single cognitive function in cognitive data to explain the mechanisms of computational language models. However, these cognitive data are the best resources we can get from the brain, and the proposed evaluation methods used in this paper could be directly extended to other cognitive data if advanced measure technology appeared in the future. 

In addition, the proposed MulCogBench only includes cognitive data in Chinese and English. Although the evaluation method is not language specific, there could be different conclusions given the varying properties of languages. Moreover, to show the effectiveness of the dataset, we take four representative language models as examples. More recent language models may give different conclusions about the relation with human cognitive data.


\bibliographystyle{lrec-coling2024-natbib}


\end{document}